\documentclass[times,twocolumn,final,authoryear]{elsarticle}

\usepackage{ycviu}
\usepackage{framed,multirow}

\usepackage{mathrsfs}
\usepackage{enumitem}
\usepackage{latexsym}
\usepackage{amssymb}
\usepackage{amsthm}
\usepackage{amsmath}
\makeatletter 
\def\@makecaption#1#2{%
  \vskip\abovecaptionskip
  \sbox\@tempboxa{#1: #2}%
  {\bfseries #1} #2\par
  \vskip\belowcaptionskip}
\makeatother
\usepackage{booktabs}
\usepackage[colorlinks,
            linkcolor=blue,
            anchorcolor=blue,
            citecolor=blue]{hyperref}

\usepackage{url}
\usepackage{xcolor}
\definecolor{newcolor}{rgb}{.8,.349,.1}

\usepackage{graphicx}
\usepackage{subfigure}

\journal{Computer Vision and Image Understanding}

\begin{document}

\thispagestyle{empty}

\clearpage
\thispagestyle{empty}

\ifpreprint
  \vspace*{-1pc}
\else
\fi

\clearpage

\ifpreprint
  \setcounter{page}{1}
\else
  \setcounter{page}{1}
\fi

\begin{frontmatter}

\title{Few-shot Action Recognition with Implicit Temporal Alignment and Pair Similarity Optimization}

\author[1,2]{Congqi \snm{Cao}\corref{cor1}}
\cortext[cor1]{Corresponding author:
  }
\ead{congqi.cao@nwpu.edu.cn}
\author[1,3]{Yajuan \snm{Li}}
\author[4]{Qinyi \snm{Lv}}
\author[1,2]{Peng \snm{Wang}}
\author[1,2]{Yanning \snm{Zhang}}

\address[1]{National Engineering Laboratory for Integrated Aero-Space-Ground-Ocean Big Data Application Technology, Northwestern Polytechnical University, Xi'an 710129, China}
\address[2]{School of Computer Science, Northwestern Polytechnical University, Xi'an 710129, China}
\address[3]{School of Cybersecurity, Northwestern Polytechnical University, Xi'an 710129, China}
\address[4]{School of Electronics and Information, Northwestern Polytechnical University, Xi'an 710129, China}

\received{1 May 2013}
\finalform{10 May 2013}
\accepted{13 May 2013}
\availableonline{15 May 2013}
\communicated{S. Sarkar}

\begin{abstract}
Few-shot learning aims to recognize instances from novel classes with few labeled samples, which has great value in research and application.
Although there has been a lot of work in this area recently, most of the existing work is based on image classification tasks. Video-based few-shot action recognition has not been explored well and remains challenging: 1) the differences of implementation details among different papers make a fair comparison difficult; 2) the wide variations and misalignment of temporal sequences make the video-level similarity comparison difficult; 3) the scarcity of labeled data makes the optimization difficult.
To solve these problems, this paper presents 1) a specific setting to evaluate the performance of few-shot action recognition algorithms; 2) an implicit sequence-alignment algorithm for better video-level similarity comparison; 3) an advanced loss for few-shot learning to optimize pair similarity with limited data.
Specifically, we propose a novel few-shot action recognition framework that uses long short-term memory following 3D convolutional layers for sequence modeling and alignment. Circle loss is introduced to maximize the within-class similarity and minimize the between-class similarity flexibly towards a more definite convergence target. Instead of using random or ambiguous experimental settings, we set a concrete criterion analogous to the standard image-based few-shot learning setting for few-shot action recognition evaluation. Extensive experiments on two datasets demonstrate the effectiveness of our proposed method.
\end{abstract}


\end{frontmatter}


\begin{figure*}[!t]
\centering
\includegraphics[scale=.5]{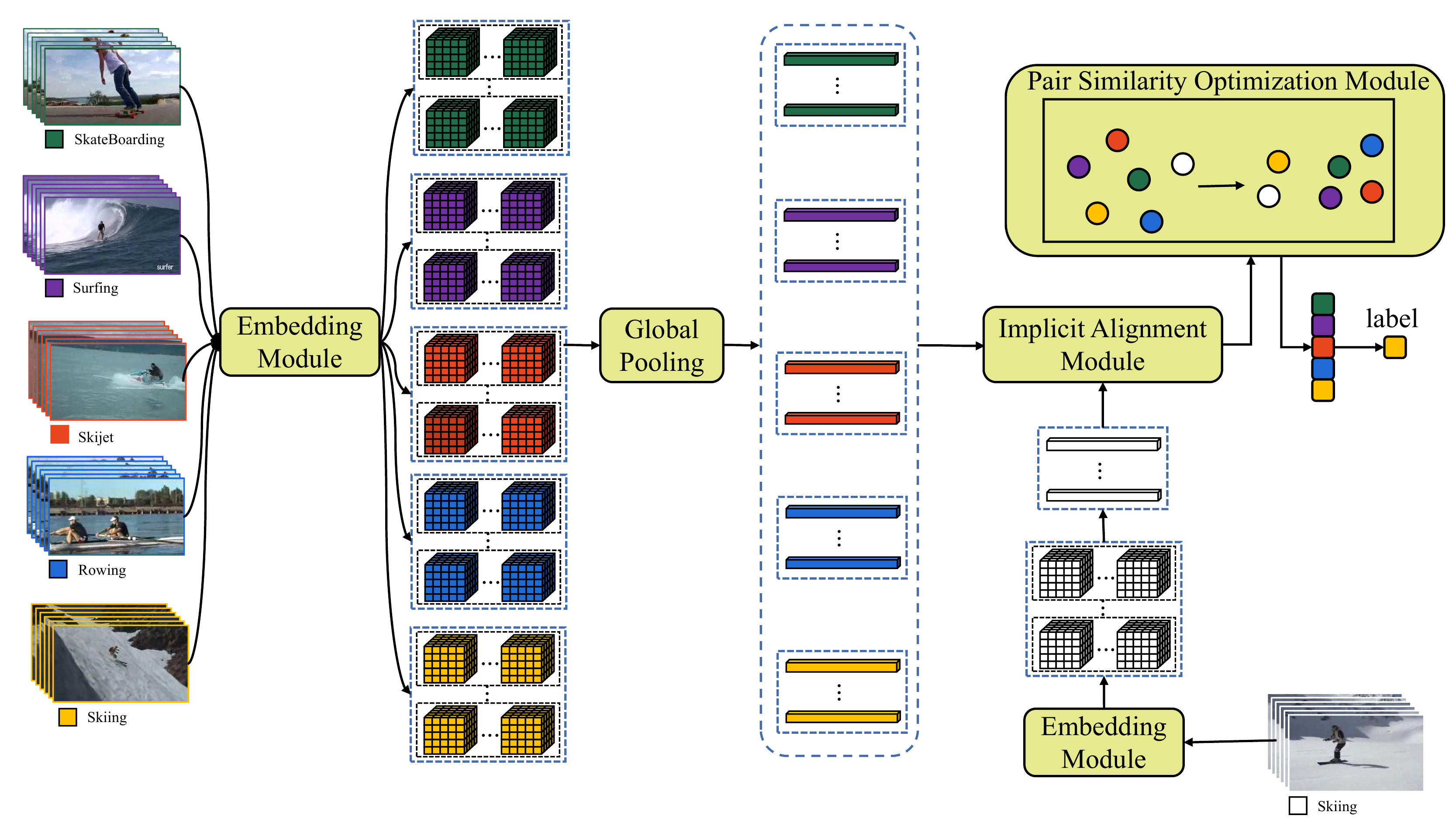}
\caption{\textbf{.} Overview of our model for few-shot action recognition in 5-way 1-shot setting. As shown, this framework consists of an embedding module, an implicit alignment module and a pair similarity optimization module. We first extract the local spatiotemporal features with the embedding module. And then the implicit alignment module is used to align a given query video with the support videos semantically at video-level. Finally, we use the pair similarity optimization module to optimize pair similarity. }
\label{proposed method}
\end{figure*}

\section{Introduction}
\label{sec1}

Few-shot action recognition aims to recognize novel actions with only a few labeled samples. It has a wide application in practical scenarios since labeling videos is expensive and there is a demand for task-specific or user-defined action recognition.
However, insufficient annotated data makes the current popular and successful deep learning-based algorithms hard to learn patterns and optimize.

Like the image-based few-shot object classification task that has been widely studied recently \citep{Compare_CVPR18, closer_iclr19, MAML_ICML17, LGM_ICML19, f_VAEGAN_CVPR19, Hallucination_CVPR19}, the solutions for video-based few-shot action recognition fall into three categories: 1) metric learning by comparing the similarity between the unknown data and the labeled samples \citep{Tan2019, Bishay2019, Cao2020}; 2) fast adaptation by learning an optimal initialization status to train a classifier \citep{Coskun2019}; 3) data augmentation by generating additional training data \citep{KumarDwivedi2019}. The last kind of solution still needs to learn a metric or train a classifier for unlabeled data recognition. The first and the second kinds of solution can be seen as learning with pair-wise labels and learning with class-level labels respectively.
As analyzed in \cite{Sun2020}, learning with class-level labels also calculates the similarity scores between samples and a set of proxies (\emph{i.e.}, classification weight vectors) representing each class. Therefore, these two learning approaches can be unified in a pair similarity optimization perspective, which aims to maximize the within-class similarity and minimize the between-class similarity.

However, it is challenging to compare and measure the similarity between videos for few-shot action recognition.
Few-shot action recognition not only faces the difficulty of data scarcity and semantic misalignment in appearance as image-based few-shot object classification does, but also needs to solve the problem of long-term temporal modeling and learning a distance measurement to compare sequences.
On images, the deformation and arbitrary location of objects may cause ambiguity when comparing two images directly.
To address this issue, \cite{SAML_ICCV19} calculate the similarity of each local region pairs of the feature maps extracted from the query image to be classified and the support images with labels to find the potential semantically relevant pairs for comparison.
\cite{Local_Descriptor_CVPR19} conduct a local descriptor based measure via k-nearest neighbor search over the deep local descriptors.
\cite{Dense_Classification_CVPR19} perform dense classification over feature maps to take full advantage of the local activations.
For action recognition, due to the additional temporal dimension, there are wide variations and enormous information in video. Directly comparing the similarity of each local region pairs in the spatiotemporal space will take a huge computation cost. Besides that, the neglect of temporal ordering information is unfavorable for action recognition since temporal information plays an important role.
To leverage the temporal information in video data, \cite{Tan2019} encode videos as dynamic images \citep{bilen2016dynamic} and use a CNN to get an incorporating feature for similarity learning.
\cite{Bishay2019} utilize a segment-by-segment attention mechanism to perform temporal alignment.
\cite{ Cao2020} explicitly align video sequences with a variant of the Dynamic Time Warping (DTW) algorithm \citep{muller2007dynamic}.
However, the routines of actions are generally not fixed since performers can change their habits, \emph{i.e.}, the location and duration of each atomic step can vary drastically, even some atomic steps can be omitted.
It is difficult to align two appearance-different sequences explicitly, especially with scarce labeled data for training.
Thus, in this paper, we propose to align video sequences implicitly and optimize video-level pair similarity directly for few-shot action recognition.

In concrete, we propose to use long short-term memory (LSTM) \citep{hochreiter1997long} following 3D convolutional embedding layers \citep{Chen2018} for sequence modeling. Instead of calculating the similarity between segment-level features, we measure the distance between video-level descriptors for semantic alignment and comparison. To alleviate the optimization challenge caused by extremely limited annotated data and find a better convergence status with a more flexible process, we adopt circle loss \citep{Sun2020}, which re-weights each similarity to highlight the less-optimized ones instead of using an equal penalty, for pair similarity optimization.
The framework of our proposed model is illustrated in Fig. \ref{proposed method}.
We design extensive experiments to evaluate the performance of our method.
Nevertheless, another outstanding issue of few-shot action recognition is lack of standard and unambiguous settings for performance evaluation. The existing work randomly divides the datasets into training, validation and test splits with different ratios \citep{Zhu2018, KumarDwivedi2019, Tan2019, Cao2020, Bishay2019, Bo2020, Zhang2020}. Even some work does not set a validation set. The number of samples selected for each training episode is inconsistent.
And the accuracies are computed by averaging over different numbers of episodes generated randomly from the test split.
To make a fair comparison with competing methods, we establish a specific and clear experimental setting on UCF101 and HMDB51 datasets, which can be used as a standard setting for few-shot action recognition evaluation.

In summary, our main contributions include:

\begin{itemize}
\item
We propose a novel few-shot action recognition model to compare video sequences with implicit alignment and long-term temporal modeling.
\item
We propose to re-weight the video-level similarity scores depending on their status flexibly during training, which benefits the optimization process.
\item
Extensive experiments are conducted with a clearly defined experimental setting that can be seen as a standard evaluation criterion for few-shot action recognition. The proposed method outperforms competing methods by 13\%-17\% in 5-way 1-shot learning and 10\%-12\% in 5-way 5-shot learning.
\end{itemize}

\section{Related Work}
\noindent
\textbf{Few-shot learning.}
Few-shot learning has attracted increasing attention recently, especially on the image-based classification task.
Current work on this task can be roughly divided into three categories: (i) metric learning-based methods, (ii) fast adaptation-based methods, and (iii) data augmentation-based methods.

The metric learning-based methods aim to learn an embedding space and a corresponding metric function to measure the similarity or distance between samples, by which intra-class samples are closer while inter-class samples are farther away \citep{siamese_ICMLW15, Prototypical_NIPS17}. For better metric learning ability, \cite{Compare_CVPR18} present a learnable relation module to compute the similarity score. \cite{adaptive_metric_NIPS18} propose a task dependent adaptive metric. \cite{huang2019compare} use pairwise bilinear pooling for feature comparison.
To alleviate the problem of object deformation and misalignment,
\cite{SAML_ICCV19} calculate the similarity scores of local region pairs and re-weight these local scores to obtain the image-level similarity.
\cite{Local_Descriptor_CVPR19} conduct k-nearest search over the local descriptors to replace the image feature based measure by a local descriptor based measure.
However, as the feature space increases, extra computational cost and parameters are introduced.
The fast adaptation-based methods aim to learn a good initialization or update strategy of a learner so that the learner can quickly adapt to new tasks.
\cite{Optimization_ICLR17} train an LSTM-based meta-learner model to learn both a good initialization and a parameter updating mechanism for the learner network.
\cite{MAML_ICML17} propose to find an initialization that can be quickly adapted to a new task via a few gradient steps.
Instead of sharing an initialization between tasks forcibly, \cite{baik2020learning} propose a task-dependent layer-wise attenuation to dynamically control how much of prior knowledge each layer would exploit for a given task.
The main problem of this kind of methods is the difficulty in applying to domains with large gap and tasks with conflicts.
Data augmentation-based methods generally supplement training samples by generative models or make full use of data characteristics through self-supervised and unsupervised learning methods \citep{f_VAEGAN_CVPR19, Hallucination_CVPR19, selfsupervision_ICCV19}.
However, this kind of methods does not solve the issue that the learning algorithms heavily depend on a large number of training samples.

\noindent
\textbf{Few-shot action recognition.}
Few-shot action recognition is a branch of few-shot learning that based on video.
Besides the spatial appearance, temporal motion information plays an important role for few-shot action recognition.

In the metric learning paradigm,
\cite{Guo2018} propose Neural Graph Matching (NGM) Network, which jointly learns a graph generator and a graph matching metric function, for 3D few-shot learning action recognition. NGM uses graphical structure to model the spatial relationship and to capture the temporal evolution of videos.
\cite{Tan2019} condense videos into images with dynamic image \citep{bilen2016dynamic}, then use a CNN to get an incorporating feature and learn a similarity metric of it.
\cite{Bo2020} propose Temporal Attention Vectors (TAVs) to encode the video-wide temporal information by calculating the weighted sum of all frame features.
Temporal Attentive Relation Network (TARN) \citep{Bishay2019} utilizes a segment-by-segment attention mechanism for temporal alignment and learns a distance measure on the aligned representations.
\cite{Cao2020} propose to use a variant of the Dynamic Time Warping (DTW) algorithm \citep{muller2007dynamic} to align and measure the distance between videos.
However, aligning the semantic content in videos is still challenging since there are wide variations and the data is extremely limited.
In the fast adaptation paradigm,
\cite{Zhu2018} introduce a multi-saliency algorithm to encode videos into matrix representations and use a compound memory network for classification.
\cite{Coskun2019} develop a low-shot transfer learning method to learn domain invariant features for first-person action recognition.
In the augmentation paradigm,
\cite{KumarDwivedi2019} propose ProtoGAN which is a conditional generative adversarial network to synthesize additional examples for novel categories.
\cite{Zhang2020} use spatial and temporal attention modules to localize actions and enrich the training data by feeding numbers of augmented clips into the network with self-supervised loss.
Generally, as the spatiotemporal modeling capacity of the model increases, the risk of overfitting also increases, since there are more parameters and few training data, which is a main problem of few-shot action recognition.

Another issue of few-shot action recognition is the lack of standard and unambiguous experimental settings for performance evaluation.
The datasets are randomly divided into training, validation, and test splits with different ratios, even without the validation split \citep{KumarDwivedi2019, Tan2019, Cao2020, Zhang2020}.
Besides that, the accuracies are computed by averaging over different numbers of episodes.
On the other side, the experimental settings in image-based few-shot classification are clearly defined, including the data splits and the evaluation criterion.

\textbf{Similarity optimization.}
The objective of few-shot learning can be seen as maximizing intra-class similarity as well as minimizing inter-class similarity \citep{ Bishay2019, Coskun2019, Cao2020}.
For metric learning-based methods, the similarity is calculated between samples. For fast adaptation-based methods with a classifier, the similarity is calculated between samples and the classification weight vectors. Thus, it is a pair similarity optimization problem.

Existing metric loss function (\textit{e.g.} triplet loss \citep{hoffer2015deep}) and classification loss function (\textit{e.g.} softmax cross-entropy loss \citep{murphy2012machine}) impose an equal penalty on each similarity score, which is inflexible and non-optimal.
\cite{Sun2020} propose to re-weight each similarity score to emphasize the less-optimized ones and verify the effectiveness of the proposed circle loss on face recognition, person re-identification, as well as fine-grained image retrieval tasks.

\section{Proposed Method}
\label{section:method}


Instead of aligning videos explicitly and rigidly as well as measuring similarity based on clip-level features \citep{Bishay2019, Cao2020}, which is non-optimal for long-term sequence modeling and hard to optimize since there are wide variations in video and the labeled data is extremely scarce, we propose a flexible video-level semantic alignment and similarity optimization mechanism for few-shot action recognition.
The framework of our proposed model is shown in Fig. \ref{proposed method}, which can be broadly divided into three major blocks: (a) an embedding module for spatiotemporal feature extraction, (b) an implicit alignment module to model long-term temporal information and align two sequences semantically in video-level, and (c) a pair similarity optimization module for adjusting gradients on similarity scores flexibly according to their optimization status.
In the following, Section \ref{formulation} describes the problem formulation of few-shot action recognition. Section \ref{embedding}, \ref{alignment}, \ref{optimization} introduce the proposed modules in detail.

\begin{figure}[!t]
\centering
\includegraphics[scale=.45]{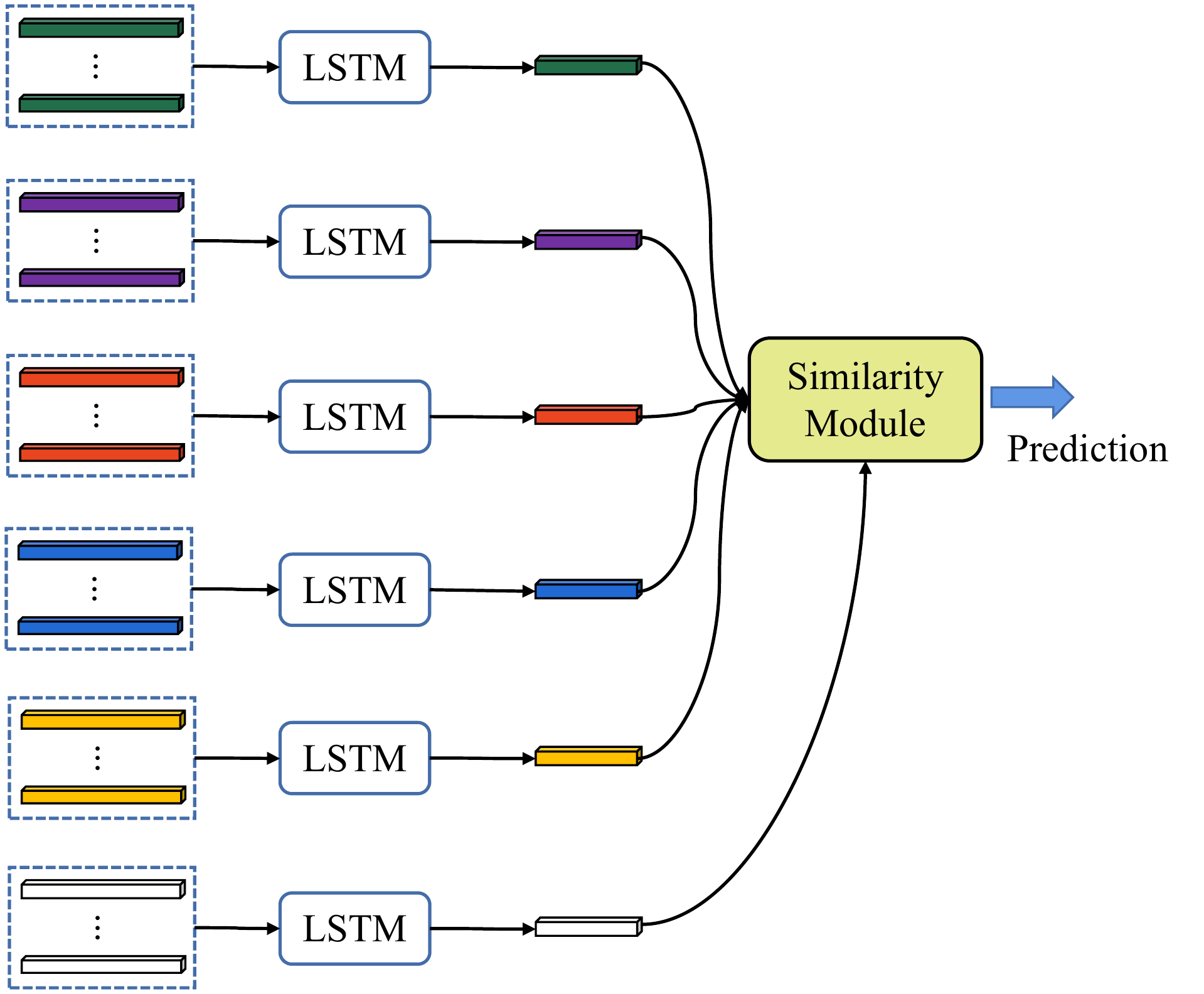}
\centering
\caption{\textbf{.} Structure of the implicit alignment module. We use long short-term memory for sequence modeling and semantic alignment. Then, the similarity module is used to compare video-level similarities.}
\label{alignment_model}
\end{figure}

\begin{figure*}[!t]
    \begin{minipage}[t]{0.45\textwidth}
       	\centering
       	\includegraphics[scale=.5]{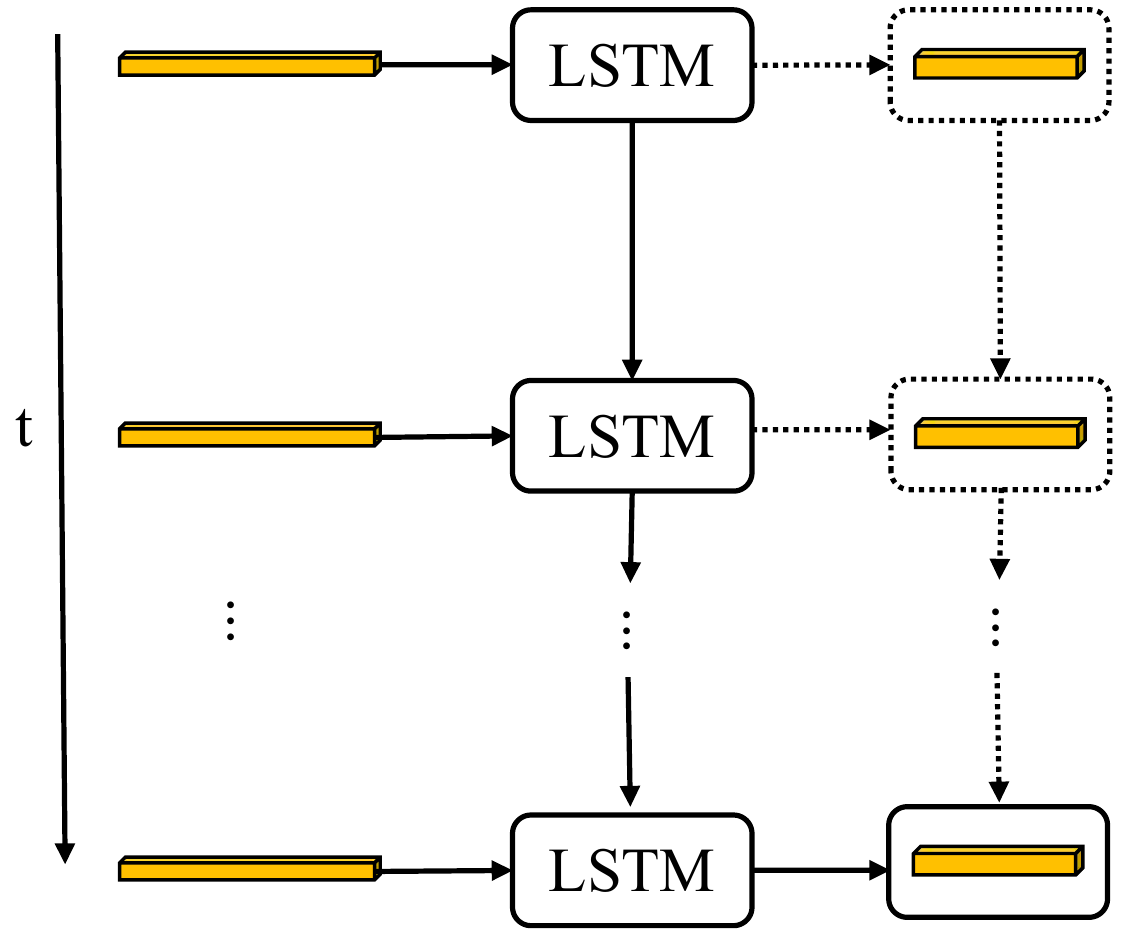}
    \end{minipage} \begin{minipage}[t]{0.5\textwidth}
   		\centering
		\includegraphics[scale=.5]{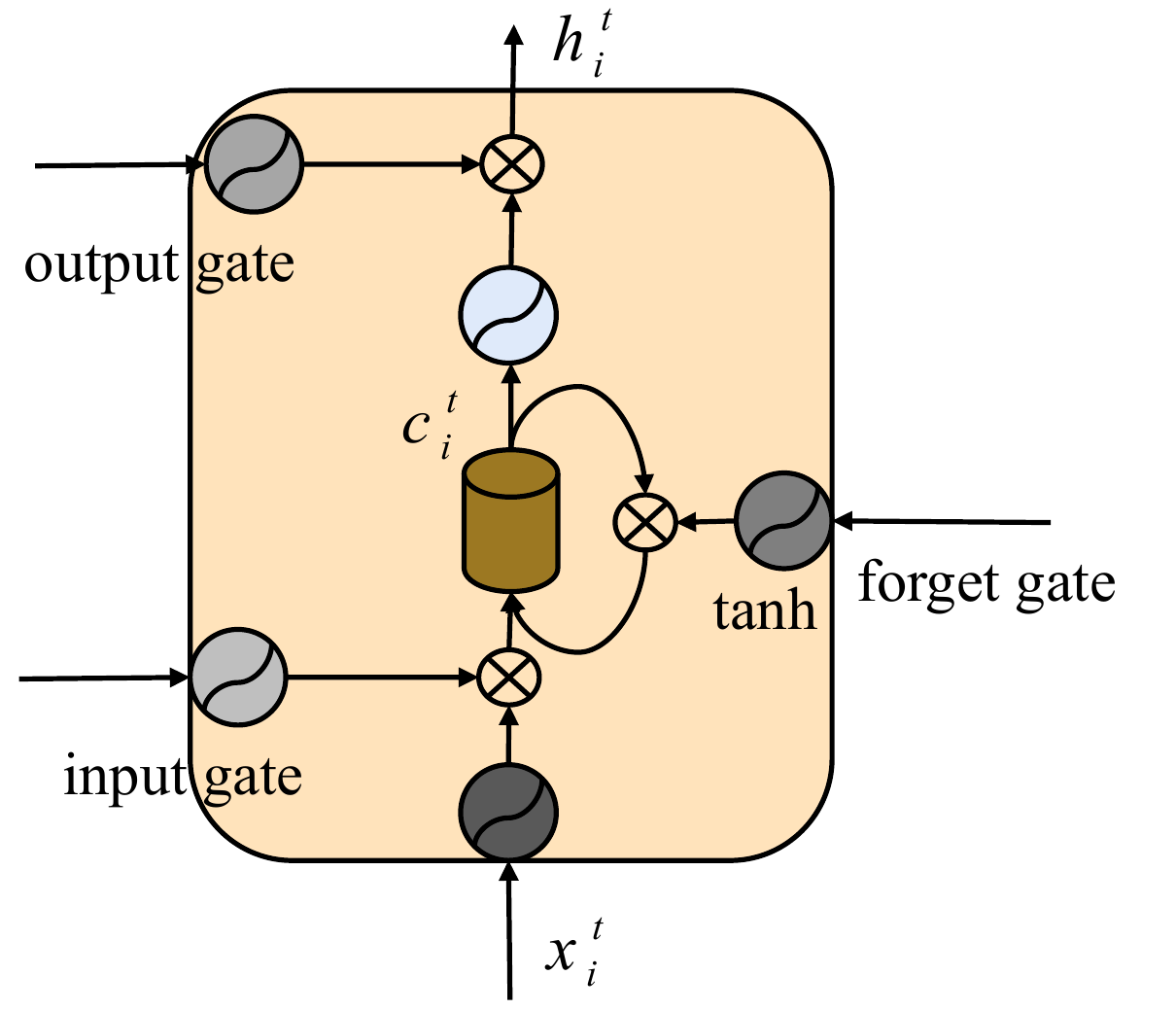}
    \end{minipage}
    \caption{\textbf{.} Illustration of sequence modeling using the long short-term memory neural network between the video clips. The structure of LSTM is shown on the right. } %
    \label{lstm}

\end{figure*}

\subsection{Problem formulation}
\label{formulation}
In few-shot action recognition setting, we divide the whole dataset into three sets: a training set $N_{train}$, a validation set $N_{validation}$, and a test set $N_{test} $. There is no overlapping category between the three sets. The training set has sufficient labeled data for each class, which minimizes the similarity loss over training episodes for few-shot learning. The validation set is used to select the optimal model. And the test set is used to evaluate the performance \textit{i.e.}, the accuracy is computed by averaging over a number of few-shot learning tasks sampled from the test set. For few-shot learning, the episode-based training method, which mimics few-shot learning tasks during training, proposed by \cite{vinyals2016matching}, is proved to be an effective training method. In the specific \textit{N}-way \textit{K}-shot few-shot learning task, each episode contains a support-set $\mathcal{S}$ sampled from the training set $N_{train}$, which contains $\textit{N}\times\textit{K}$ samples from \textit{N} different classes where each class contains \textit{K} support samples. Then \textit{Q} samples from each class are selected to form the query set $\mathcal{Q}$ which contains $\textit{N}\times\textit{Q}$ samples. The goal is to classify the \textit{N}$\times$\textit{Q} query samples into \textit{N} classes only with the \textit{N}$\times$\textit{K} support samples. In our setting, we give a specific number of the query set $\mathcal{Q}$ in each episode, which is different from the earlier methods without a clearly defined value. Moreover, there is no standard for the division ratio of $N_{train}$, $N_{validation}$ and $N_{test}$. Some work even does not have $N_{validation}$. And the categories of $N_{train}$, $N_{validation}$ and $N_{test}$ are randomly selected. However, the similarity between the categories in $N_{train}$, $N_{validation}$ and $N_{test}$ has a strong influence on learning performance. We use an unambiguous and deterministic data splitting criterion in our paper to eliminate the influence of different divisions. In addition, the accuracies are computed by averaging over different numbers of episodes in the existing work. We give a specific and detailed evaluation setting in our experiments as a reference for subsequent work to make a fair comparison.

\subsection{Embedding Module}
\label{embedding}
The embedding module $f_{\varphi }$ generates the feature representations for the query and support videos, where $ f $ represents the projection function with learnable parameters $ \varphi $.
Discriminative spatiotemporal information is important while challenging to extract for few-shot action recognition with few labeled training data. Therefore, we choose to use a light-weight 3D convolutional neural network with good performance in action recognition to capture the spatiotemporal information. In our paper, the convolutional layers of the multi-fiber network (MFNet) \citep{kay2017kinetics} pre-trained on Kinetics-400 is selected as the embedding module $f_{\varphi }$ to extract video features.

We divide a video into \textit{T} clips. Each clip has a fixed number of frames.
For an input video sequence $ x=\lbrace x^{1}, x^{2}, ..., x^{T}\rbrace$, we encode each clip $x^{t}$ ($t \in [1, ..., T]$) with the embedding module into feature representation $f_{\varphi}(x^{t})$, which makes $ f_{\varphi}(x) $ a set of feature vectors:
\begin{equation}
f_{\varphi}(x)=\lbrace f_{\varphi}(x^{1}), f_{\varphi}(x^{2}), ..., f_{\varphi}(x^{T}) \rbrace
\end{equation}
where the embedded feature dimension of each clip $x^{t}$ is $(C, {T}',H,W)$. Specifically, $ C $ indicates the channel dimension, $ {T}' $ represents the temporal dimension, $ H $  and $ W $ are the height and width, respectively.

\subsection{Implicit Alignment Module.}
\label{alignment}
The implicit alignment module  $ \mathcal{A}_{\psi} $ aims to model long-term temporal information and align two sequences semantically in the video-level. The structure of this module is illustrated in Fig. \ref{alignment_model}. First, we use long short-term memory for sequence modeling and implicit semantical alignment, as shown in Fig. \ref{lstm}. Next, the similarity module is used to optimize the video-level similarities.

In one episode, each support video $ x_{i} $ from the support set $ \mathcal{S} =\lbrace (x_{i}, y_{i}) \rbrace_{i=1}^{G} $ $ (G=N \times K, y_{i}\in \lbrace 1,...,N\rbrace)$, and the query video $q$ from the query-set $ \mathcal{Q} $ are projected into the embedding feature space by the embedding module $f_{\varphi} $, which is followed by a dimensionality reduction function $ \mathcal{R} $ as the following equation:

\begin{equation}\label{eq_2}
\mathcal{R}(f(x_{i}))=\lbrace \mathcal{R}(f_{\varphi}(x_{i}^{1})), \mathcal{R}(f_{\varphi}(x_{i}^{2})), ..., \mathcal{R}(f_{\varphi}(x_{i}^{T}))\rbrace
\end{equation}
where $ \mathcal{R} $ is a spatiotemporal dimension reduction function such as a global average/max pooling layer to prevent overfitting. We use global average pooling in our experiment.

Then, we use the long shot-term memory (LSTM) network to model the temporal relationship of the feature sequence $(\mathcal{R}(f_{\varphi}(x_{i}^{1})), \mathcal{R}(f_{\varphi}(x_{i}^{2})), ..., \mathcal{R}(f_{\varphi}(x_{i}^{T})))$ and align two sequences implicitly, as shown in Fig. \ref{lstm}. The LSTM computes the sequence of outputs $(\mathcal{R}(f_{\varphi}(x_{i}^{1})), \mathcal{R}(f_{\varphi}(x_{i}^{2})), ..., \mathcal{R}(f_{\varphi}(x_{i}^{T})))$  by iterating the following equations:
\begin{equation}
I_{i}^{t}=\sigma(W_{xI}f_{\varphi}(\mathcal{R}(x_{i}^{t}))+W_{hI}h_{i}^{t-1}+W_{cI}\circ c_{i}^{t-1}+b_{I})
\end{equation}
\begin{equation}\label{3}
f_{i}^{t}=\sigma(W_{xf}f_{\varphi}(\mathcal{R}(x_{i}^{t}))+W_{hf}h_{i}^{t-1}+W_{cf}\circ c_{i}^{t-1}+b_{f})
\end{equation}
\begin{equation}\label{4}
c_{i}^{t}=f_{t}\circ c_{i}^{t-1}+I_{i}^{t}\circ tanh(W_{xc}f_{\varphi}(\mathcal{R}(x_{i}^{t}))+W_{hc}h_{i}^{t-1}+b_{c})
\end{equation}
\begin{equation}\label{5}
o_{i}^{t}=\sigma(W_{xo}f_{\varphi}\mathcal{R}((x_{i}^{t}))+W_{ho}h_{i}^{t-1}+W_{co}\circ c_{i}^{t}+b_{o})
\end{equation}
\begin{equation}\label{6}
h_{i}^{t}=o_{i}^{t}+tanh(c_{i}^{t})
\end{equation}
where $ I_{i}^{t} $, $ f_{i}^{t} $,  $ c_{i}^{t} $, $ o_{i}^{t} $, $ h_{i}^{t} $ indicate input gate, forget gate, cell status, output gate, hidden layer output respectively, and $\circ$ denotes the Hadamard product, and $t\in[1,T]$. We use the output of the hidden layer $ h_{i}^{t} $ of the last time slice $T$ as a video-level feature for $ x_{i} $, which is denoted as $ \mathcal{A}_{\psi}(x_{i})$. For the query video $ q $ in each episode, we can get its video-level feature $ \mathcal{A}_{\psi}(q) $ as above.

Next, we calculate the similarity between the video-level features $ \mathcal{A}_{\psi}(x_{i}) $ and $ \mathcal{A}_{\psi}(q) $ by:
\begin{equation}
cos(\mathcal{A}_{\psi}(x_{i}),\mathcal{A}_{\psi}(q))=\dfrac{\mathcal{A}_{\psi}(x_{i})^{\intercal} \mathcal{A}_{\psi}(q)}{\parallel \mathcal{A}_{\psi}(x_{i})^{\intercal} \parallel\cdot \parallel \mathcal{A}_{\psi}(q) \parallel}
\end{equation}
where $cos(\cdot)$ indicates the cosine similarity whose value range is [-1, 1], and $\parallel \cdot \parallel$ means L2 norm. Other similarity or distance functions can also be employed.

The sequences are implicitly aligned by maximizing the similarity scores between the samples belonging to the same class and minimizing the similarity scores between the samples belonging to different classes during few-shot learning.
For K-shot where $K>1$, we average over the embedding features of all the samples from the same class to form a prototype feature of this class. The class probability of the query video is computed based on the similarity between the query video and the prototype features of all the classes.

\subsection{Pair Similarity Optimization Module}
\label{optimization}
Few-shot learning can be seen as maximizing intra-class similarity as well as minimizing inter-class similarity. Given a query video $ q $, we calculate the similarity scores between $ q $ and the videos of the support set $ \mathcal{S} $ in each episode. For the negative sample from the inter-class, we denote its similarity score with the query video by $ s_{n}^{j} = \dfrac{\mathcal{A}_{\psi}(x_{j})^{\intercal} \mathcal{A}_{\psi}(q)}{\parallel \mathcal{A}_{\psi}(x_{j})^{\intercal} \parallel \cdot \parallel \mathcal{A}_{\psi}(q) \parallel } $ ($ x_{j} $ is the \textit{j}-th sample in the negative sample set $ \mathcal{N} $ of the support set  $ \mathcal{S} $ ).
And for the positive sample from the intra-class, we denote its similarity score with the query video by $ s_{p}^{i} = \dfrac{\mathcal{A}_{\psi}(x_{i})^{\intercal} \mathcal{A}_{\psi}q)}{\parallel \mathcal{A}_{\psi}(x_{i})^{\intercal} \parallel \cdot \parallel \mathcal{A}_{\psi}(q) \parallel } $ ($ x_{i} $ is the \textit{i}-th sample in the positive sample set $ \mathcal{P} $ of the support set  $ \mathcal{S} $).
Maximizing the intra-class similarity as well as minimizing the inter-class similarity is equivalent to minimizing $ s_{n}-s_{p} $ by loss function:
\begin{equation}\label{eq_10}
\mathcal{L}
= log [1+\sum_{i=1}^{L} \sum_{j=1}^{M} exp(\gamma (s_{n}^{j}-s_{p}^{i}+m))]
\end{equation}
where $ L=\mid \mathcal{P} \mid $, $ M=\mid\mathcal{N}\mid $, $ \gamma $ is a scale factor and $m$ is a margin for better similarity separation.

However, the above method imposes an equal penalty on each similarity score, which is inflexible and non-optimal. Therefore, we follow recent work \citep{Sun2020} to generalize $ s_{n}^{j}- s_{p}^{i} $ into $\alpha_{n}^{j} s_{n}^{j}- \alpha_{p}^{i} s_{p}^{i} $, which allows each similarity score to learn at its own pace. The loss function without margin is shown in Eq. \ref{eq_11}. Note that $ \alpha_{n} $ and $ \alpha_{p} $ are independent weighting factors, allowing $s_{n} $ and $ s_{p} $ to learn at different paces.

\begin{equation}\label{eq_11}
\begin{aligned}
\mathcal{L}_{circle}
&= log [1+\sum_{i=1}^{L} \sum_{j=1}^{M} exp(\gamma  (\alpha_{n}^{j} s_{n}^{j}- \alpha_{p}^{i} s_{p}^{i}))]\\
&=log [1+\sum_{j=1}^{M} exp(\gamma \alpha_{n}^{j} s_{n}^{j}) \sum_{i=1}^{L} exp(- \gamma \alpha_{p}^{i} s_{p}^{i})]
\end{aligned}
\end{equation}
in which $ \alpha_{n}^{j} $ and $ \alpha_{p}^{i} $ are non-negative weighting factors.

Assume that the optimal value of $ s_{p}^{i} $ is $ O_{p} $ and the optimal value of $ s_{n}^{j} $ is $ O_{n} $. When a similarity score deviates far from its optimum, it should get a large weighting factor so as to get effective update with large gradient. We follow \cite{Sun2020} to define $ \alpha_{p}^{i} $ and $ \alpha_{n}^{j} $ as follows:
\begin{equation}\label{eq_12}
\begin{cases}
\alpha _{p}^{i}=\left [ O_{p}-s_{p}^{i} \right ]_{+}
 \\
 \\
\alpha _{n}^{j}=\left [ s_{n}^{j}-O_{n} \right ]_{+}
\end{cases}
\end{equation}
in which $ [\cdot]_{+} $ is the ``cut-off at zero'' operation to ensure $ \alpha_{p}^{i} $ and $ \alpha_{n}^{j} $ are non-negative.

Since $ s_{n} $ and $ s_{p} $ are in asymmetric position in Eq. \ref{eq_11}. Naturally, it requires respective margins for $ s_{n} $ and $ s_{p} $, which is formulated by:
\begin{small}
\begin{equation}\label{eq_13}
\mathcal{L}_{circle}= log [1+\sum_{j=1}^{M} exp(\gamma \alpha_{n}^{j}(s_{n}^{j}-\triangle _{n})) \sum_{i=1}^{L}exp(-\gamma \alpha_{p}^{i}(s_{p}^{i}-\triangle _{p}))]
\end{equation}
\end{small}
in which $ \triangle _{n}$ and $ \triangle _{p}$ are the inter-class and intra-class margins, respectively.

Eq. \ref{eq_13} expects $s_{p}^{i} > \triangle _{p}$ and $s_{n}^{j} < \triangle _{n}$.
The setting of $\triangle _{p}$ and $\triangle _{n}$ can be analyzed by deriving the decision boundary.
For simplicity, the case of binary classification is considered, in which the decision boundary is achieved at $ \alpha_{n} (s_{n}-\triangle _{n})-\alpha_{p} (s_{p}-\triangle _{p})=0 $. With Eq. \ref{eq_12} and Eq. \ref{eq_13}, the decision boundary is formulated as:
\begin{equation}\label{eq_14}
(s_{n}-\frac{O_{n}+\triangle_{n}}{2})^{2}+ (s_{p}-\frac{O_{p}+\triangle_{p}}{2})^{2}=C
\end{equation}
in which $ C= ((O_{n}-\triangle_{n})^{2} + (O_{p}-\triangle_{p})^{2})/4 $. Eq. \ref{eq_14} shows that the decision boundary is the arc of a circle, so Eq. \ref{eq_13} is named circle loss.

There are five hyper-parameters for circle loss, \textit{ i.e.}, $ O_{p} $, $ O_{n} $ in Eq. \ref{eq_12} and $ \gamma $, $ \triangle_{p} $, $ \triangle_{n} $ in Eq. \ref{eq_13}. The hyper parameters can be reduced by setting $ O_{p}=1+m $, $ O_{n}=-m $, $ \triangle _{n}=m $ and $ \triangle _{p}=1-m$. Hence there are only two hyper-parameters, \textit{i.e.}, the scale factor $ \gamma $ and the relaxation margin \textit{m}.

The gradients of circle loss with respect to $ s_{n}^{j} $ and $ s_{p}^{i} $ are derived as follows:
\begin{equation}\label{eq_15}
\frac{\partial \mathcal{L}_{circle}}{\partial s_{n}^{j}}=Z \frac{exp ( \gamma ((s_{n}^{j})^{2}-m^{2}))}{\sum_{l=1}^{M}exp (\gamma ((s_{n}^{l})^{2}-m^{2}))} \gamma (s_{n}^{j}+m)
\end{equation}
 and
\begin{equation}\label{eq_16}
\frac{\partial \mathcal{L}_{circle}}{\partial s_{p}^{i}}=Z \frac{exp ( \gamma (m^{2}-(s_{p}^{i}-1)^{2}))}{\sum_{k=1}^{L}exp (\gamma (m^{2}-(s_{p}^{k}-1)^{2}))} \gamma (s_{p}^{i}-1-m)
\end{equation}
in both of which $ Z=1-exp(-\mathcal{L}_{circle}) $

In our experiments, we verify the efficiency and effectiveness of re-weighting the similarity scores in the episode-based training stage.
It is more flexible and can achieve better performance compared with traditional loss functions using equal penalty on each similarity score.

\begin{figure*}[!t]
\centering
\includegraphics[scale=.5]{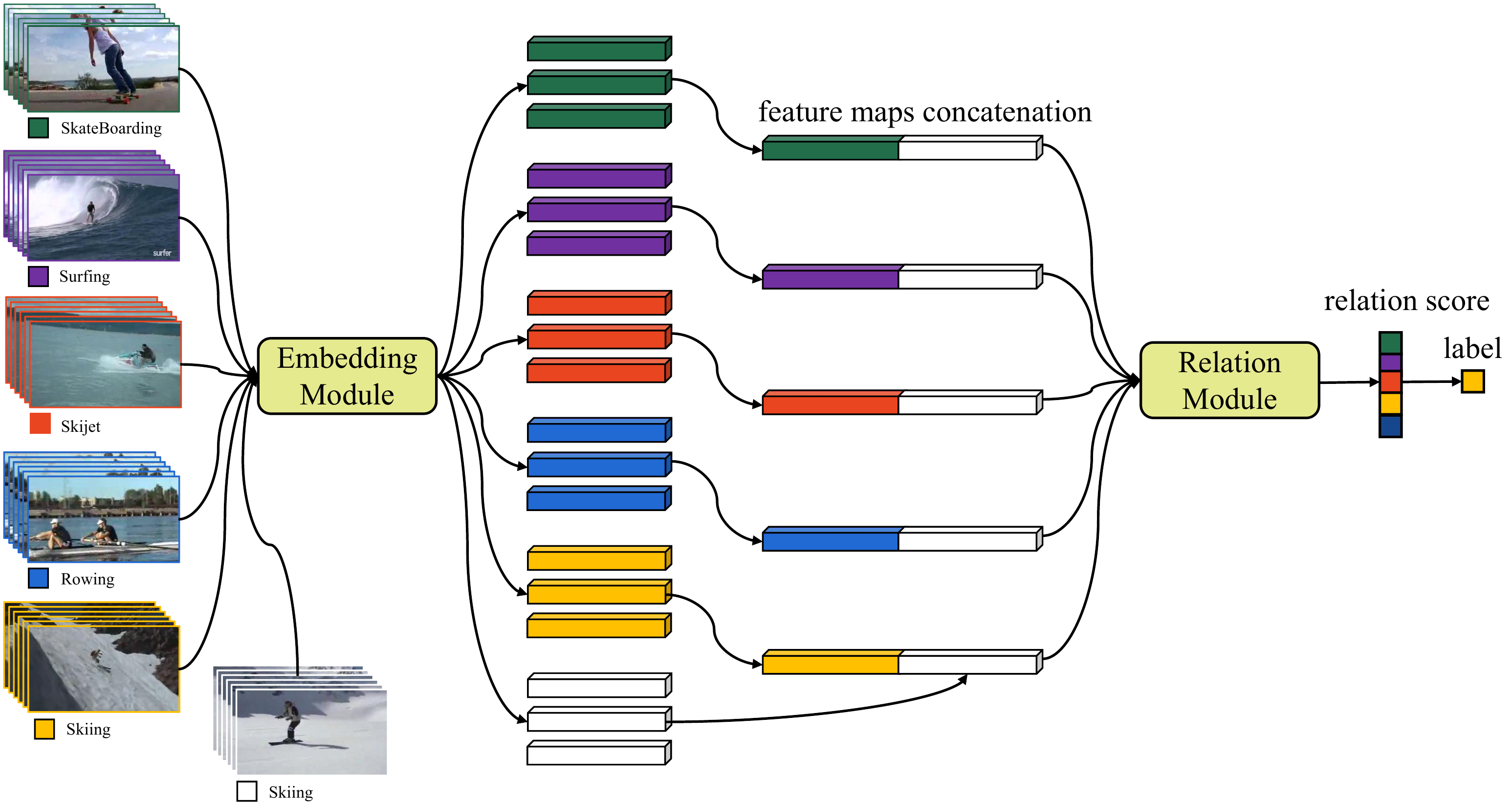}
\caption{\textbf{.} Framework of the baseline Relation Video Network (RVN). It first extracts video features using the embedding module. Then the mean feature is computed as the centroid of each class. Next, the embedding features of the query video and each support video are concatenated. Finally, the relation module is used to compute the relation score between the query video and videos in the support set.}
\label{baseline}
\end{figure*}

\section{Experiments}
In this section, we conduct comprehensive experiments to evaluate the proposed few-shot action recognition model and compare it with extensive baselines. The datasets used for evaluation are described in Section \ref{dataset}. Implementation details of the training and testing mechanism are described in Section \ref{implementation}. Detailed experimental results are described in section \ref{experimental_result}.

\subsection{Datasets}
\label{dataset}
We use two action recognition datasets to evaluate models, \textit{i.e.,} the UCF101 dataset and the HMDB51 dataset.

\textbf{UCF101} \citep{Soomro2012} is an action recognition dataset of realistic action videos collected from YouTube. There are 101 action categories and 13,320 videos. UCF101 is very challenging for few-shot action recognition which exists largest diversity in terms of actions and large variations in camera motion. For our experiments, we follow the splitting setting used by \cite{Zhang2020} which chooses 70, 10 and 21 non-overlapping classes for training, validation and testing, respectively. The specific action classes in the training split, validation split, and test split are listed in the supplementary material.

\textbf{HMDB51} \citep{Kuehne2011} collects from various sources, mostly from movies, and a small proportion from public databases such as the Prelinger archive, YouTube and Google videos.
It contains 6,849 videos divided into 51 action categories. Each category contains at least 101 videos. For our experiments, we follow the splitting setting used by \cite{Zhang2020} which chooses 31, 10 and 10 non-overlapping classes for training, validation, and testing, respectively. The specific action classes used for training, validation, and test are listed in the supplementary material.

\subsection{Implementation details}
\label{implementation}
In an \textit{N}-way \textit{K}-shot task setting, we sample $ N\times K $ videos from \textit{N} classes where each class has \textit{K} examples as the support set. To be specific, 5-way 1-shot and 5-way 5-shot classification tasks are conducted on all datasets. During training, we sample 18,000 episodes to train our model.
In each episode, we select \textit{N} samples to build the query set, where each sample in the query set belongs to one of the \textit{N} classes in the support set. During testing, we select 15 unlabeled samples from each class of \textit{N} as the query set. Thus each testing episode has a total of \textit{N} (\textit{K} + 15) examples. We sample 600 episodes from the test split in the experiments. This process is repeated ten times, and the final mean accuracy is reported. Moreover, the 95\% confidence intervals are also reported.
For both datasets, we divide each video into $ T $ segments, where $ T=3 $ in our experiment. In each segment, we uniformly sample 16 frames to form a clip. We use the input size of  $3 \times 16 \times 112\times 112$ for training and testing. Here 3, 16, and $ 112 \times 112 $ are the channel dimension, the number of frames for each input clip and the spatial resolution of each frame, respectively. We follow the video preprocessing procedure in multi-fiber network (MFNet) \citep{Chen2018}. During training, we first resize the frames in the sampled clip to $ 256 \times 256  $ and then randomly crop a $ 112 \times 112 $ area. In the inference phase, we change the random crop to center crop.
We apply stochastic gradient descent \citep{bottou2010large} to optimize our model during the episode-based few-shot learning stage, with an initial learning rate of 0.00001 and decaying every 12,000 episodes by 0.1. All experiments are conducted using PyTorch \citep{Paszke2017}.

\subsection{Experimental Results}
\label{experimental_result}
In this section, we first describe the architecture of baseline model in Section \ref{baseline_method}. Then, we presents ablation studies in Section \ref{ablation}. We report the results of our proposed framework compared with other methods in Section \ref{result}. Finally, we visualize the similarity matrices in Section \ref{visual}.

\begin{figure}[!t]
\centering
\includegraphics[scale=.5]{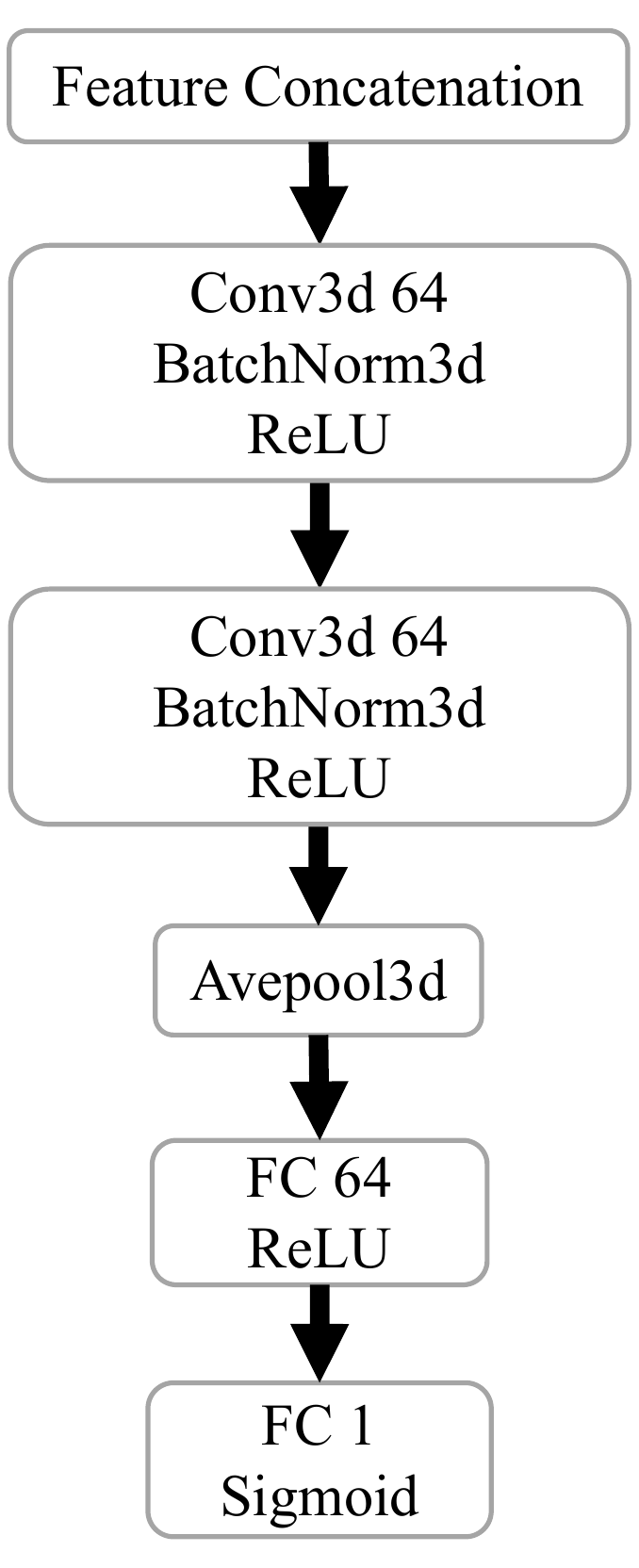}
\caption{Structure of the relation module in RVN.}
\label{realtion}
\end{figure}

\subsubsection{Baseline model}
\label{baseline_method}
To verify the importance of modeling spatiotemporal relationships, we design a baseline architecture, Relation Video Network (RVN). Our baseline model RVN consists of two modules: an embedding module $f_{\varphi }$ and a relation module $g_{\phi}$, as shown in Fig. \ref{baseline}.

In each episode, the support video $x_{i}$ from the support-set $\mathcal{S}$ and the query video $\emph{q}$ from the query-set $\mathcal{Q}$ are fed into the embedding module $f_{\varphi }$, which generates the embedding feature maps $ f_{\varphi}(x_{i})=\lbrace f_{\varphi}(x_{i}^{1}), f_{\varphi}(x_{i}^{2}), ..., f_{\varphi}(x_{i}^{T}) \rbrace$ and $ f_{\varphi}(q)=\lbrace f_{\varphi}(q^{1}), f_{\varphi}(q^{2}), ..., f_{\varphi}(q^{T}) \rbrace$, respectively. The video-level features of the support video and the query video are obtained by averaging over the $T$ clip-level feature maps, which are denoted as $\mathcal{M}(f_{\varphi}(x_{i}))$ and $ \mathcal{M}(f_{\varphi}(q))$, respectively, where $ \mathcal{M}$ represents the mean operation. Then, the video-level features are concatenated by the operator of $\mathcal{C}\left (\mathcal{M}(f_{\varphi}(x_{i})), \mathcal{M}(f_{\varphi}(q)) \right )$. Subsequently, the combined feature of the support video and query video is fed into a relation module $g_{\phi}$, which generates a relation score between 0-1 to indicate the similarity of the support video $x_{i}$ and the query video $\emph{q}$. The structure of the relation module $g_{\phi}$ is shown in Fig. \ref{realtion}.
Specifically, it mainly contains two convolution layers, a global average pooling layer, and two fully-connected layers. Each convolution layer is composed of a 3D convolution with 64 filters, a batch normalization and a ReLU activation.
The dimensions of the two fully-connected layers are 64 and 1, respectively. Finally, a sigmoid activation is used to obtain the relation score in [0, 1] range.

\subsubsection{Ablation Studies}
\label{ablation}
Here we perform ablation experiments to verify the effectiveness of modeling spatiotemporal relationships, verify the importance of long-term temporal modeling and explore the impact of circle loss. The default setting is 5-way 1-shot learning.

\begin{table}[!t]
\setlength{\abovecaptionskip}{0pt}
\setlength{\belowcaptionskip}{10pt}
\caption{\newline The 5-way 1-shot accuracy results of different structures of the relation module on UCF101 dataset. The second column indicates the size of the convolution kernels used by the relation module in RVN.}
\label{table1}
\begin{tabular*}{\hsize}{@{}@{\extracolsep{\fill}}lllllllllllll@{}}
\toprule
Model       & \multicolumn{2}{c}{Convolution kernel}                              & \multicolumn{5}{c}{Accuracy (\%)}                              \\ \midrule

RVN\--1 & \multicolumn{2}{c}{$ 3\times 3 \times 3 $, $  3\times 3 \times 3 $}                            &                          \multicolumn{5}{c}{\textbf{80.33 $ \pm $ 0.25 }} &                            &                            \\
RVN\--2 &              \multicolumn{2}{c}{$ 3\times 3 \times 3 $, $ 3\times 1 \times 1 $}               &            \multicolumn{5}{c}{77.83 $ \pm $ 0.26
}
                &                            &                            \\
RVN\--3 &          \multicolumn{2}{c}{$ 1\times 3 \times 3 $, $ 1\times 3 \times 3 $}                  &       \multicolumn{5}{c}{76.87 $ \pm $ 0.27}
        &                            &                            \\
\bottomrule
\end{tabular*}
\end{table}

\begin{table}[!t]
\setlength{\abovecaptionskip}{0pt}
\setlength{\belowcaptionskip}{10pt}
\caption{\newline  The 5-way 1-shot accuracy results of different LSTM structures on UCF101 dataset.}
\label{table2}
\begin{tabular*}{\hsize}{@{}@{\extracolsep{\fill}}lllllllllllll@{}}
\toprule
 \multicolumn{1}{c}{Units in hidden layer}  &  \multicolumn{4}{c}{Accuracy (\%)}                                              \\ \midrule
 \multicolumn{1}{c}{512}        &     \multicolumn{4}{c}{\textbf{86.57 $ \pm $ 0.21}}           \\
\multicolumn{1}{c}{256}  		&    \multicolumn{4}{c}{85.14 $ \pm $ 0.22}          				\\
\multicolumn{1}{c}{64}			&    \multicolumn{4}{c}{74.74 $ \pm $ 0.25}          			\\
\bottomrule
\end{tabular*}
\end{table}

\textbf{Influence of relation.} To verify the effectiveness of modeling spatiotemporal relationships, we perform an ablation study by developing three different relation modules and analyzing the effects of different relation modules. Specifically, the first relation module named RVN\--1 uses $ 3 \times 3 \times 3 $ convolutional kernels in both convolution layers. As for the second relation module, RVN\--2, it performs a spatiotemporal convolution with $ 3 \times 3 \times 3 $  kernels and a temporal convolution with $ 3 \times 1 \times 1 $ kernels.  And the third relation module named RVN\--3 uses $ 1 \times 3 \times 3 $ convolution kernels in both convolution layers which only does spatial convolution.
Under the 5-way 1-shot setting, the results on the UCF101 dataset are listed in Table \ref{table1}.
The accuracy of RVN\--1 gains 2.5\% and 3.46\% improvements over RVN\--2 and RVN\--3 respectively for the 5-way 1-shot learning task.
The results reflect the importance of modeling spatiotemporal information simultaneously.

\textbf{Different structures of LSTM.} In RVN, the video-level features of the support videos and query videos are obtained by averaging over the \textit{T} clip-level feature maps, in which temporal ordering information is lost. Therefore, we propose to use long shot-term memory to model the long-term temporal information of \textit{T} clips. We explore the influence of different structures of LSTM in Table \ref{table2}.
As mentioned in Section \ref{embedding}, the dimension of the embedded clip-level features is $(C, {T}', H, W)$. Using the convolutional layers of MFNet as the embedding layer, $ C $, $ {T}'$, $ H $, and $ W $ equal 768, 8, 4, and 4, respectively.
To prevent overfitting, we use global average pooling to reduce the spatiotemporal dimension from (8, 4, 4) to (1, 1, 1) before feeding forward into LSTM. Therefore, the input dimension of LSTM is \textit{C}=768. The corresponding results of different numbers of the units in hidden layer are listed in Table \ref{table2}. Relatively, high dimensional features are more effective for sequence modeling and semantic alignment when comparing video-level similarities in few-shot action recognition.
In addition, the results in Table \ref{table2} show that long-term temporal information is essential for few-shot action recognition compared with the results in Table \ref{table1}.

\begin{table}[!t]
\setlength{\abovecaptionskip}{0pt}
\setlength{\belowcaptionskip}{10pt}
\caption{\newline Evaluation of cross-entropy loss and circle loss. We report the 5-way 1-shot on UCF101 dataset.  }
\label{table3}
\begin{tabular*}{\hsize}{@{}@{\extracolsep{\fill}}lllllllllllll@{}}
\toprule

Model                           & \multicolumn{1}{c}{Loss Function} & \multicolumn{2}{c}{Accuracy (\%)}        \\
  \midrule
 Our method       & \multicolumn{1}{c}{Circle loss}  & \multicolumn{2}{c}{\textbf{88.71 $ \pm $ 0.19}}
              \\
Our method         & \multicolumn{1}{c}{Cross-entropy  loss}   & \multicolumn{2}{c}{86.57 $ \pm $ 0.21} \\
   \bottomrule
\end{tabular*}
\end{table}

\begin{figure}[!t]
	\centering
    \subfigure[Cross-entropy loss.]{
    \begin{minipage}[t]{0.45\textwidth}
    	\centering   	    	
       	\includegraphics[scale=.5]{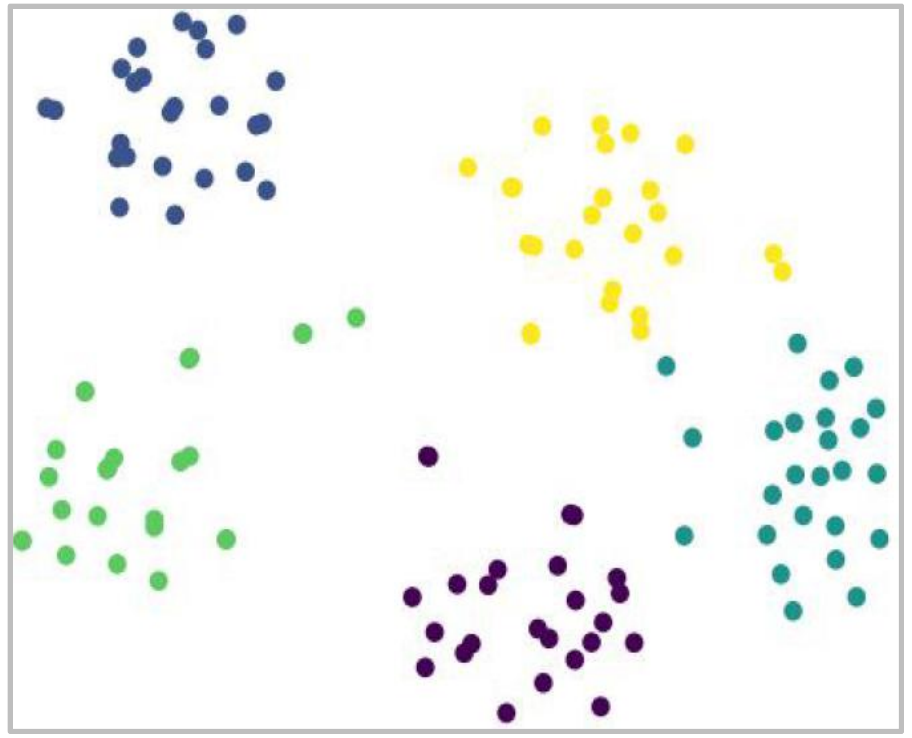}       	
    \end{minipage} %
    } \subfigure[Circle loss.]{
	\begin{minipage}[t]{0.45\textwidth}		
		\centering
		\includegraphics[scale=.5]{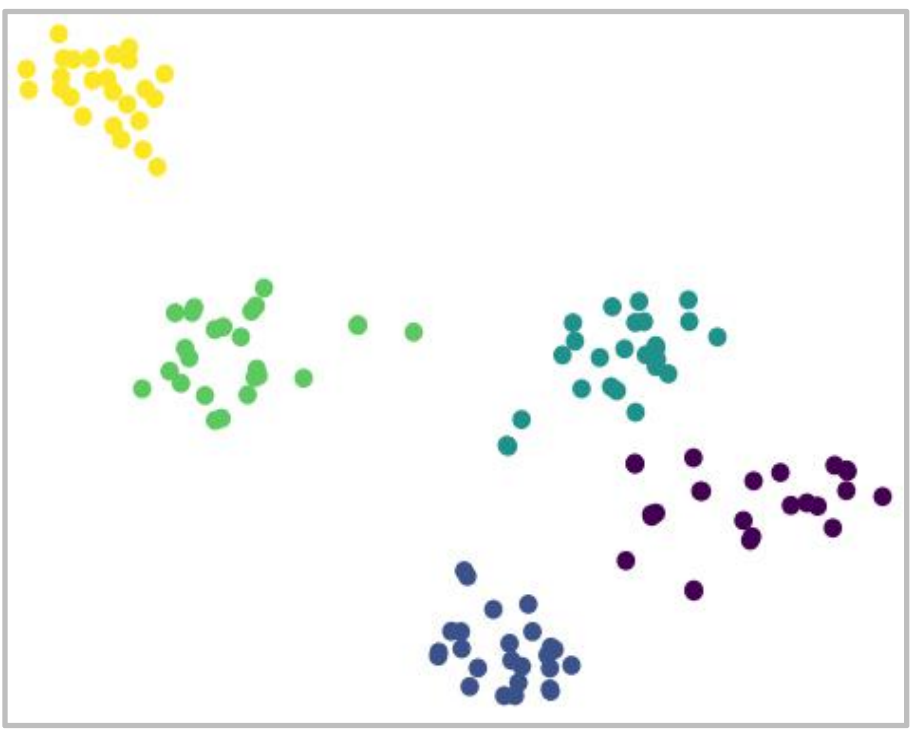}
    \end{minipage} %
    }
    \caption{\textbf{.}  The t-SNE visualization \citep{maaten2008visualizing} of the $ \mathcal{A}(\cdot) $ feature. (a) corresponds to the 5-way 1-shot setting of our model using CEL and (b) corresponds to our model using the circle loss. We can see that circle loss  makes clusters more compact and discriminative from each other.} %
    \label{loss}
\end{figure}

\begin{figure}[!t]
	\centering
    \subfigure[scale factor $\gamma$.]{
    \begin{minipage}[t]{0.45\textwidth}
    	\centering   	    	
       	\includegraphics[scale=.5]{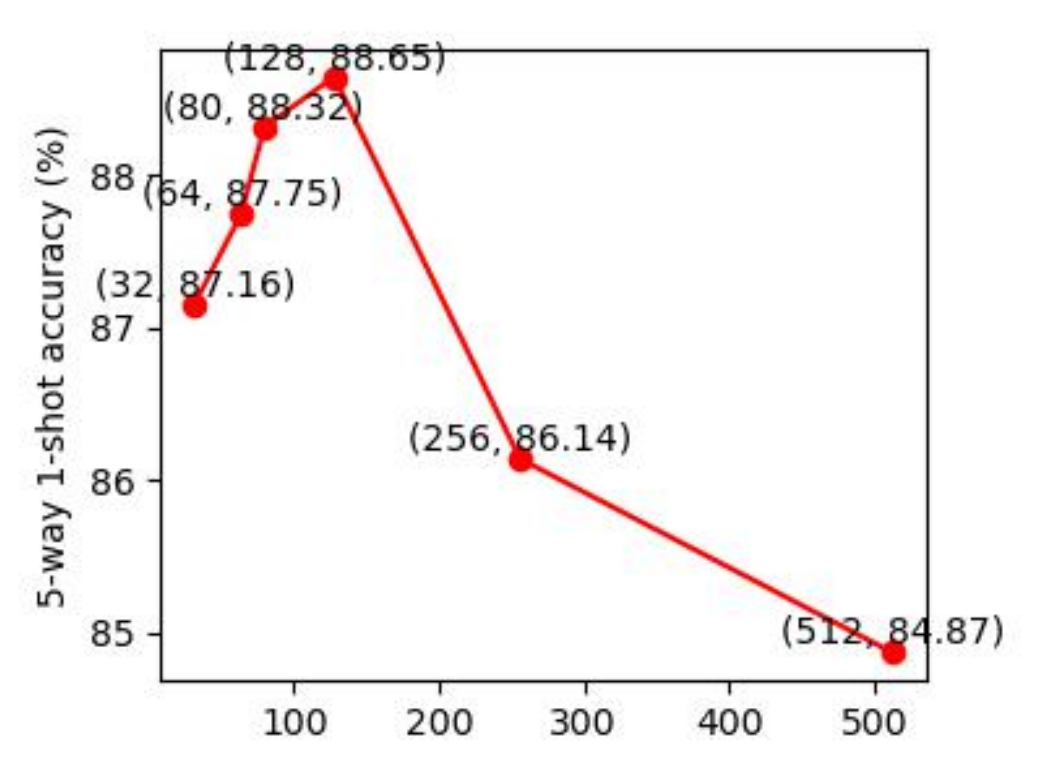}       	
    \end{minipage} %
    } \subfigure[relaxation factor \textit{m}.]{
	\begin{minipage}[t]{0.45\textwidth}		
		\centering
		\includegraphics[scale=.5]{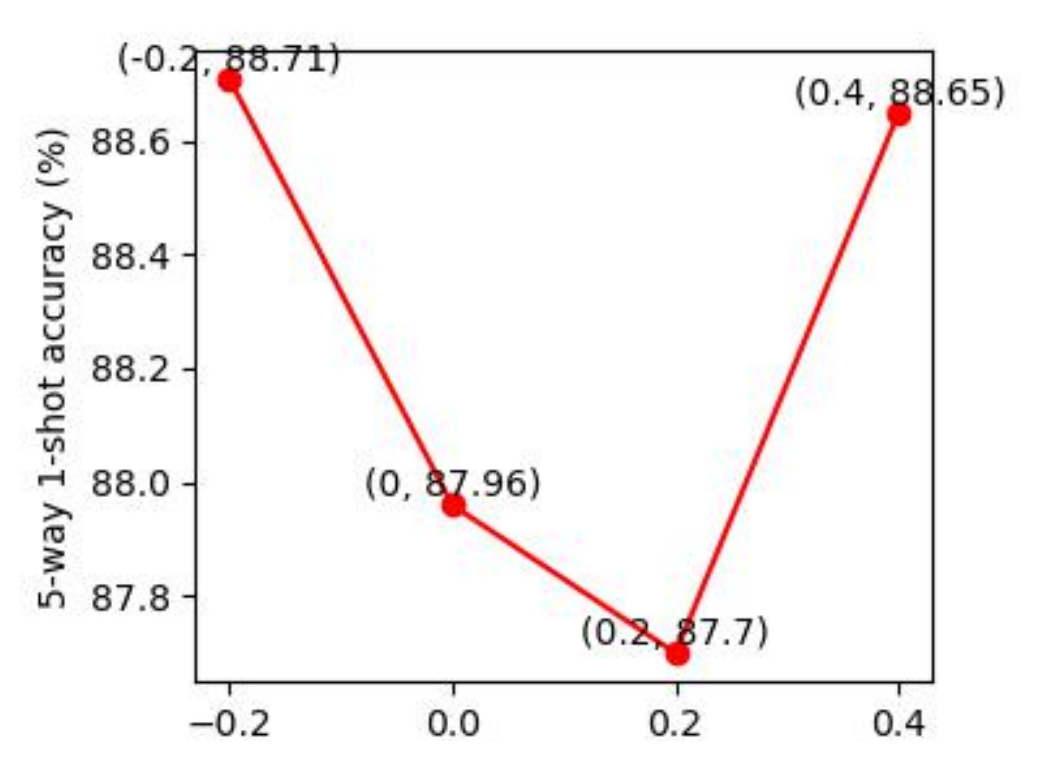}
    \end{minipage} %
    }
    \caption{\textbf{.} Impact of the two hyperparameters. (a) shows the accuracies corresponding to different values of the scale factor $ \gamma $, and (b) shows the accuracies corresponding to different values of the relaxation factor \textit{m} with 5-way 1-shot setting on UCF101 dataset.} %
    \label{gamma_m}
\end{figure}

\begin{table*}[!t]
\setlength{\abovecaptionskip}{0pt}
\setlength{\belowcaptionskip}{10pt}
\caption{\newline Few-shot action recognition results. The mean accuracies of the 5-way 1-shot and 5-shot tasks on UCF101 and HMDB51 datasets are listed, with 95\% confidence intervals. }
\label{table4}
\begin{tabular*}{\hsize}{@{}@{\extracolsep{\fill}}lllllllllllll@{}}
\toprule
\multicolumn{1}{l}{}                           & \multicolumn{4}{c}{5-Way Accuracy (\%)}                                              \\
Model       & \multicolumn{2}{c}{UCF101}          & \multicolumn{2}{c}{HMDB51}          \\
                                             & \multicolumn{1}{c}{1-shot} & 5-shot & \multicolumn{1}{c}{1-shot} & 5-shot \\ \midrule

Generative Approach  \citep{Rai}                           & \multicolumn{1}{l}{-}	& 78.68 $ \pm $ 1.80	&	-	&	52.58 $ \pm $ 3.10
               \\
Feature-Aligning Network (FAN) \citep{Tan2019}                         & \multicolumn{1}{l}{71.80 $ \pm $ 0.10 }	&	86.50 $ \pm $ 0.20	&	50.20 $ \pm $ 0.20	&	67.60 $ \pm $ 0.10
				\\
ProtoGAN  \citep{KumarDwivedi2019}                        &
\multicolumn{1}{l}{61.70 $ \pm $ 1.60}	&	79.70 $ \pm $ 0.80	&	34.40	 $ \pm $ 1.30&	50.90 $ \pm $ 0.60
				\\
Action Relation Network (ARN) \citep{Zhang2020}          &
\multicolumn{1}{l}{62.10 $ \pm $ 1.00}	&	84.80 $ \pm $ 0.80	&	44.60	$ \pm $ 0.90	&	59.10  $ \pm $ 0.80

\\ \midrule
Our Method                            &                                   \multicolumn{1}{l}{\textbf{88.71 $ \pm $ 0.19}}	&	\textbf{96.78 $ \pm $ 0.08}	&	\textbf{63.43 $ \pm $ 0.28}	&	\textbf{79.69 $ \pm $ 0.20}
              \\   \bottomrule
\end{tabular*}
\end{table*}

\begin{figure*}[!t]
	\centering
	\subfigure[Similarity matrix learned by our model.]{
    \begin{minipage}[t]{0.45\textwidth}
    	\centering		
		\includegraphics[scale=.5]{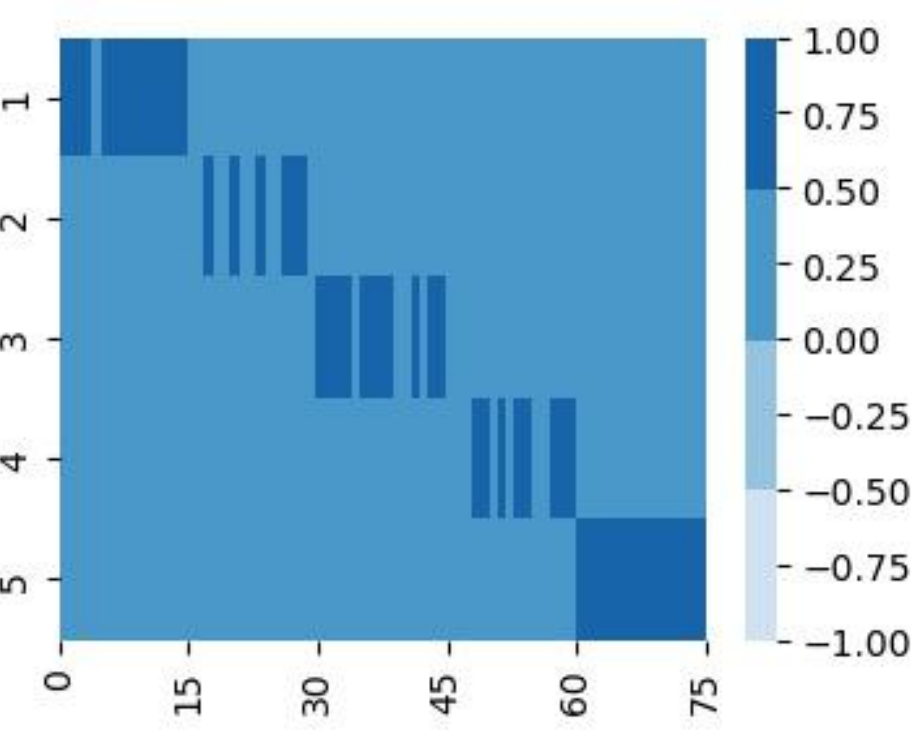}
    \end{minipage}
    }
    \subfigure[Ground truth of our model.]{
    \begin{minipage}[t]{0.45\textwidth}   		
    	\centering	
		\includegraphics[scale=.5]{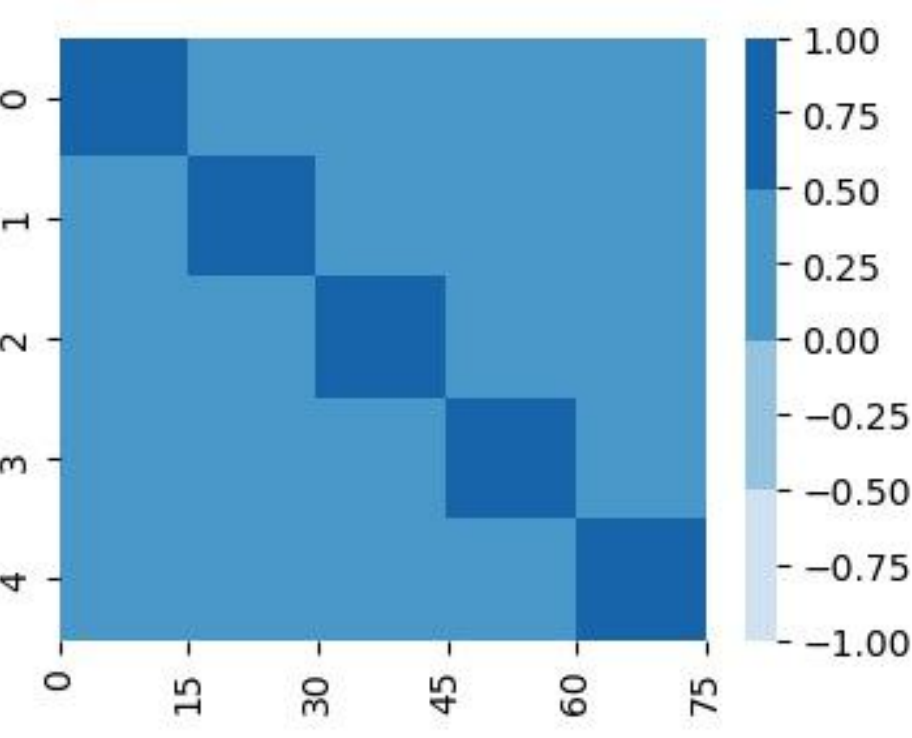}
    \end{minipage}
    }
    \subfigure[Similarity matrix learned by RVN\--1.]{
	\begin{minipage}[t]{0.45\textwidth}		
		\centering	
		\includegraphics[scale=.5]{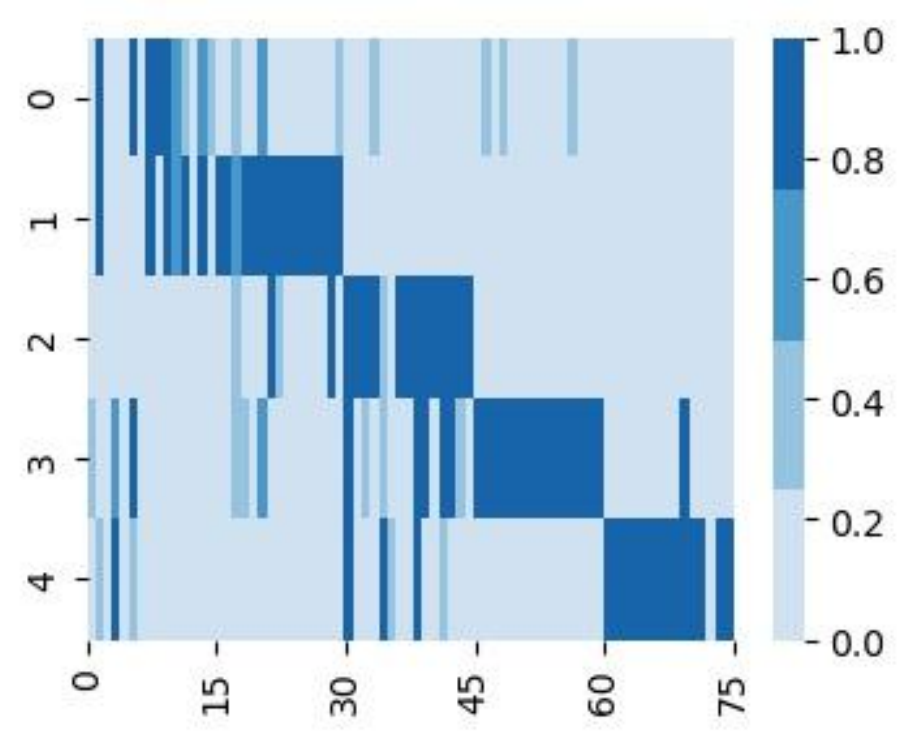}
    \end{minipage} %
    }
    \subfigure[Ground truth of RVN\--1.]{
    \begin{minipage}[t]{0.45\textwidth}
    	\centering	  	    	
       	\includegraphics[scale=.5]{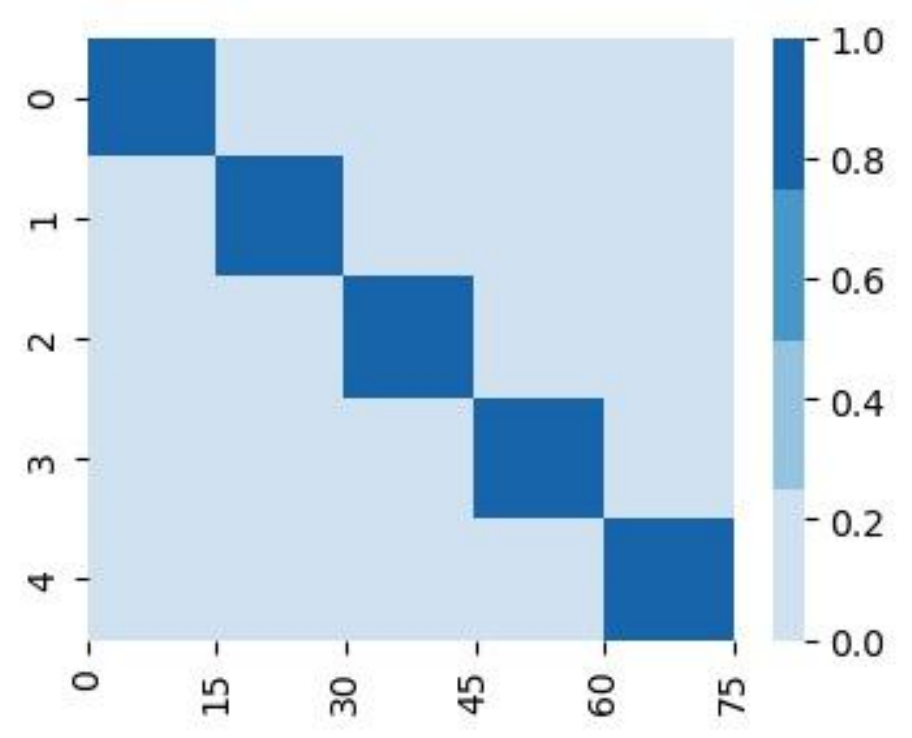}       	
    \end{minipage} %
    } %
    \caption{\textbf{.} Similarity matrices learned by our proposed model,  the baseline RVN\--1, and the ground truth on UCF101 dataset under the 5-way 1-shot learning setting. Vertical axis denotes the index of the five classes in the support set. Horizontal axis denotes the index of the $ 15 \times 5 $ query videos.} %
    \label{vis}
\end{figure*}

\textbf{Influence of circle loss.}
In the pair similarity optimization module, we use circle loss to optimize our model. In order to prove that circle loss is a more effective optimization method, we compare it with cross-entropy loss (CEL) in this part. The performance with 5-way 1-shot setting on the UCF101 dataset is shown in Table \ref{table3}. When using circle loss to optimize our model, the accuracy reaches 88.71 $ \pm $ 0.19 \%, which is 2.14\% higher and with a more compact confidence interval than that of CEL. As can be seen, with the circle loss, our model can perform better than the case of using the CEL.
In addition, in order to observe the effects of CEL and circle loss on the features, we visualize the feature distribution of 5-way 1-shot setting using t-SNE \citep{maaten2008visualizing} in Fig. \ref{loss}. Compared with CEL, circle loss makes clusters more compact and discriminative from each other.
In Fig. \ref{gamma_m}, we analyze the impact of the two hyperparameters for circle loss, \textit{i.e.}, the scale factor $ \gamma $ and the relaxation factor \textit{m}. We set $ \gamma $ to be 32, 64, 80, 128, 256, and 512 in Fig. \ref{gamma_m} (a) with a fixed value of $m$, where $m=0.4$. As can be seen, when $ \gamma $ varies between 32 and 128, circle loss surpasses CEL and has good robustness.
Then we set $ \gamma $ to be 128 and vary \textit{m} from -0.2 to 0.4 (with 0.2 as the interval) and visualize the results in Fig. \ref{gamma_m} (b). Under these settings, circle loss surpasses the performance of CEL (86.57$ \pm $ 0.21\%) consistently.

\subsubsection{Evaluating Few-Shot Learning}
\label{result}
We compare our proposed approach with other few-shot action recognition methods in this section.
By default, we conduct 5-way few-shot action recognition. The 1-shot and 5-shot action recognition results on both the UCF101 and HMDB51 datasets are listed in Table \ref{table4}. It can be observed that our method compares favorably against other methods by a large margin, not limited to the metric-based methods but also including the augmentation-based methods.
Compared with ARN \citep{Zhang2020}, the performance is 88.71\% vs 62.10\%,  96.78\% vs. 84.80\%, 63.43\% vs. 44.60\%, and 79.69\% vs. 59.10\% under 5-way 1-shot and 5-way 5-shot settings on the two benchmarks UCF101 and HMDB51, respectively.
Compared with FAN \citep{Tan2019}, our method gains 16.91\%, 10.28\%, 13.23\%, and 12.09\% improvements on 5-way 1-shot and 5-way 5-shot action recognition tasks on UCF101 and HMDB51 datasets, respectively.
By comparing with the competing methods, we could conclude that considering long-term temporal information, aligning video sequences implicitly, and allowing each similarity score to learn at its own pace help to enhance the generalization ability of the model on unseen classes for few-shot action recognition.

\subsubsection{Visualization}
\label{visual}
We visualize the similarity matrices learned by our proposed model with the  implicit alignment and pair similarity optimization modules, and the baseline RVN\--1 under the 5-way 1-shot learning setting on UCF101 dataset. We select 15 query videos from each class (\emph{i,e.}, 75 query videos in total), calculate the similarity between each query video and the support video in each class, and visualize the $ 5\times75 $ similarity matrices. Note that the ground truth similarity scores for circle loss are different from those for CEL as introduced in Section \ref{optimization}.
From Fig. \ref{vis}, it can be seen that the similarity scores learned by our proposed model are much closer to the ground truth than those learned by the baseline model, which demonstrates the superior learning ability of our propose modules for few-shot action recognition.

\section{Conclusion}
In this paper, we propose a novel few-shot action recognition model. It aligns video sequences implicitly with long short-term memory units following 3D convolutional layers for sequence comparison, and adjusts the gradients on the video-level similarity scores dynamically during training for a more flexible and better optimization process. In addition, we present a standard and unambiguous experimental setting for performance evaluation.
Extensive experiments verify the superiority of our method, which can handle the problem of semantic misalignment and wide variation in videos well with extremely limited data and achieve state-of-the-art accuracy on few-shot action recognition.



\section*{Declaration of Competing Interest}
The authors declare that they have no known competing financial interests or personal relationships that could have appeared to influence the work reported in this paper.

\section*{Acknowledgments}
This work is supported by the National Natural Science Foundation of China under Grant 61906155 and U19B2037, the Natural Science Foundation of Shaanxi Province under Grant 2020JQ-216, the China Postdoctoral Science Foundation under Grant 2020M673488, the Fundamental Research Funds for the Central Universities under Grant 31020180QD138, and the Open Projects Program of National Laboratory of Pattern Recognition.

\bibliographystyle{model2-names}
\bibliography{refs}

\section*{Supplementary Material}
\subsection*{The splitting setting of the UCF101 dataset.}
\textit{Training classes of the UCF101 data:} MilitaryParade, JavelinThrow, Biking, Drumming, SoccerJuggling, Haircut, Typing, FrontCrawl, FrisbeeCatch, Kayaking, WritingOnBoard, LongJump, BandMarching, HulaHoop, CricketBowling, PullUps, JumpingJack, BodyWeightSquats, Mixing, Basketball, Rafting, ShavingBeard, RopeClimbing, Hammering, Fencing, Archery, TrampolineJumping, ApplyEyeMakeup, BalanceBeam, TableTennisShot, Lunges, PlayingCello, JugglingBalls, HeadMassage, Knitting, BasketballDunk, Swing, PizzaTossing, WallPushups, BlowDryHair, TaiChi, Billiards, FieldHockeyPenalty, Nunchucks, BabyCrawling, Skijet, SumoWrestling, BrushingTeeth, PoleVault, BoxingPunchingBag, PlayingViolin, YoYo, BenchPress, PlayingDhol, UnevenBars, Rowing, Bowling, MoppingFloor, PushUps, SoccerPenalty, PlayingFlute, PlayingTabla, BaseballPitch, PlayingPiano, WalkingWithDog, PlayingSitar, BreastStroke, BoxingSpeedBag, ParallelBars, ThrowDiscus.

\textit{Validation classes of the UCF101 data:} ApplyLipstick, HammerThrow, Shotput, HighJump, HandstandPushups, SkateBoarding, CricketShot, HorseRiding, PlayingDaf, PlayingGuitar.

\textit{Test classes of the UCF101 data:} BlowingCandles, RockClimbingIndoor, PommelHorse, HorseRace, Skiing, HandstandWalking, JumpRope, CleanAndJerk, CliffDiving, VolleyballSpiking, Surfing, Diving, StillRings, SalsaSpin, FloorGymnastics, Punch, GolfSwing, IceDancing, CuttingInKitchen, TennisSwing, SkyDiving.

\subsection*{The splitting setting of the HMDB51 dataset.}
\textit{Training classes of the HMDB51 data:} shoot bow, climb, hug, catch, clap, pullup, wave, throw, dribble,  ride bike , sword, kiss, fall floor, punch, jump, chew, shake hands , situp,  brush hair , stand, turn, dive,  flic flac , drink, handstand,  climb stairs, sword exercise, ride horse, draw sword, push, walk.

\textit{Validation classes of the HMDB51 data:} laugh, golf, smile, cartwheel, swing baseball, somersault, shoot gun, shoot ball, hit, eat.

\textit{Test classes of the HMDB51 data:} pushup, kick ball, pick, run, kick, talk, sit, smoke, fencing, pour.
\end{document}